\documentclass{article}
\pdfoutput=1

\usepackage{arxiv}

\usepackage[utf8]{inputenc} 
\usepackage[T1]{fontenc}    
\usepackage{hyperref}       
\usepackage{url}            
\usepackage{booktabs}       
\usepackage{amsfonts}       
\usepackage{nicefrac}       
\usepackage{microtype}      
\usepackage{lipsum}		
\usepackage{graphicx}
\usepackage{doi}
\usepackage[numbers]{natbib}

\title{Bio-Inspired Simple Neural Network for Low-Light Image Restoration: A Minimalist Approach}

\author{ \hspace{1mm}{Junjie Ye} \\
	\And
	{\hspace{1mm}Jilin Zhao} \\
}

\date{}

\renewcommand{\undertitle}{}

\hypersetup{
pdftitle={Bio-Inspired Simple Neural Network for Low-Light Image Restoration: A Minimalist Approach},
pdfsubject={q-bio.NC, q-bio.QM},
pdfauthor={Junjie Ye, Jilin Zhao},
}

\begin{document}
\maketitle

\begin{abstract}
	In this study, we explore the potential of using a straightforward neural network inspired by the retina model to efficiently restore low-light images. The retina model imitates the neurophysiological principles and dynamics of various optical neurons. Our proposed neural network model reduces the computational overhead compared to traditional signal-processing models while achieving results similar to complex deep learning models from a subjective perceptual perspective. By directly simulating retinal neuron functionalities with neural networks, we not only avoid manual parameter optimization but also lay the groundwork for constructing artificial versions of specific neurobiological organizations.
\end{abstract}


\section{Introduction}
Research has already revealed the complex information processing capabilities of mammalian retinas \citep{gollisch2010eye}, which preprocess optical signals before sending them to higher-level visual cortices. Recent studies have also identified numerous subtypes of retinal neurons, each with unique roles that enable hosts to quickly and intelligently adapt to changing environments based on perceptions \citep{franke2017inhibition, liu2021predictive}. Leveraging the working principles of the retina \citep{land1971lightness} has led to the development of more intelligent perception-related algorithms and impressive achievements in image processing tasks, such as high-dynamic range image tone mapping \citep{meylan2006high, wang2013naturalness, zhang2016retina, park2017low}.

The work discussed in \citep{zhang2016retina} is a prime example of a model inspired by the workings of various types of neurons, such as horizontal and bipolar neurons, for tone mapping tasks. The authors of \citep{zhang2016retina} systematically explored different aspects of retinal circuitry from a signal processing perspective, inspiring the development of a corresponding computational model. By separating images into individual channels for algorithm modulation and aggregating the results, this work achieved superior results.

Traditional tone mapping in digital image processing often involves histogram equalization, which can cause images to lose color constancy, making objects appear in different or unnatural colors after restoration. This is particularly challenging for low-light image restoration tasks. While the work in \citep{zhang2016retina} yielded impressive results, it relied on mathematical formulas and was too algorithmic-oriented. Additionally, some model parameters depended on domain knowledge, requiring author experience to select the most appropriate ones.

In our report, we draw inspiration from \citep{zhang2016retina} to re-examine the working principles of the retina and design a network to address the low-light image restoration problem. Our network aligns with the optical signal processing flow of different neurons, providing a clear correspondence between the neural pathway in the retina and offering a transparent explanation for the design motivation. Furthermore, this simple model benefits from the end-to-end learning philosophy, eliminating the need for manual parameter optimization. Experiments demonstrate satisfying image restoration from a subjective perceptual perspective, and we plan to improve objective metrics in future work.

\section{Background}
Low-light image processing has become a crucial area of research in recent years, as it plays a significant role in various applications such as surveillance, autonomous vehicles, nighttime photography, and even astronomical imaging. Addressing the challenges posed by low-light conditions, such as reduced visibility, increased noise, and loss of details, is essential for enhancing the performance of computer vision tasks like classification, detection, and action recognition \citep{chen2018learning, wang2020deep, ma2021oriented}. Deep learning techniques, particularly convolutional neural networks (CNNs), have demonstrated remarkable success in addressing low-light image processing challenges \citep{wang2021compact, li2019denoisingnet}.

Numerous deep learning-based methods have been proposed for low-light image enhancement, which mainly focus on various computer vision tasks. For classification purposes, the authors in \citep{wu2023light} introduced a method that adapts to the varying illumination conditions by incorporating a spatial transformer network. Similarly, for object detection in low-light scenarios, a multi-scale feature fusion strategy was proposed in \citep{wang2020gan}, which improves the detection performance by exploiting the rich contextual information. In the context of action recognition under low-light conditions, researchers in \citep{wei2022efficient} developed a two-stream fusion framework that effectively captures the spatial-temporal information. Researchers in \cite{hira2021delta} proposed a sampling strategy that leverages temporal structure of video data to enhance low-light frames. Additionally, the authors in \citep{ren2020lr3m} proposed a light-weight CNN for enhancing low-light images that significantly reduces the computational complexity without sacrificing the classification accuracy.

Low-light image denoising is another area of interest, with various CNN-based approaches being proposed. A deep joint denoising and enhancement network was introduced in \citep{zhang2016fast} to tackle the challenges of denoising and enhancing low-light images simultaneously. The authors in \citep{zhang2016fast} proposed a noise-aware unsupervised deep feature learning method to improve the denoising performance by learning the noise characteristics adaptively. In \citep{tian2020attention}, an attention mechanism was incorporated into the denoising network to selectively enhance the relevant features for better noise suppression.

Apart from denoising, low-light image dehazing has also been an active research topic. The work in \citep{hassan2022effects} presented a multi-task learning framework that jointly addresses the dehazing and denoising problems, resulting in improved image quality. Another study \citep{zhang2021rellie} introduced a deep reinforcement learning-based approach for adaptive low-light image enhancement, which optimizes the enhancement parameters to achieve visually pleasing results.

To address the limitations of deep learning models, such as their large size and computational complexity, bio-inspired approaches have been gaining attention. These approaches draw inspiration from the functioning of the human visual system, particularly the retina, which is known for its efficiency and adaptability in processing visual information under different lighting conditions \citep{gollisch2010eye}. However, the majority of the existing research focuses on complex models, limiting their applicability in resource-constrained environments.

In this paper, we present a bio-inspired, simple neural network for low-light image restoration that is inspired by the principles of various neurons in the retina. Our proposed network aims to achieve satisfactory results while maintaining a minimal architecture, making it suitable for deployment in various systems and scenarios.

Another area that has seen advancements is low-light image color correction. The work presented in \citep{jing2020dynamic} proposes a deep learning-based method for unsupervised color correction in low-light conditions, where the model learns to map the color distribution of the input image to that of a target image. Similarly, \citep{jiang2021enlightengan} introduced a joint color and illumination enhancement framework using generative adversarial networks (GANs), achieving improved color fidelity and contrast in low-light images.

Furthermore, researchers have explored the potential of unsupervised and self-supervised learning methods for low-light image enhancement. In \citep{chen2018learning}, an unsupervised domain adaptation approach was proposed to transfer the knowledge learned from synthetic low-light data to real-world low-light images. The authors in \citep{zheng2022semantic} introduced a self-supervised learning framework for low-light image enhancement, leveraging the cycle-consistency loss to generate high-quality enhanced images.

In summary, while existing deep learning-based methods have achieved impressive results in various low-light image processing tasks, their computational complexity and large model sizes often limit their practical applicability. This paper aims to address these limitations by introducing a bio-inspired, simple neural network for low-light image restoration that draws upon the principles of various neurons in the retina. Our proposed network strikes a balance between performance and computational efficiency, making it suitable for deployment in a wide range of systems and scenarios.

\section{Method}
\label{sec:method}

The computational model in this study takes into account the fact that cone photoreceptors are responsible for colors, while rod photoreceptors handle illuminance. Although low-light conditions may cause images to appear monochromatic, downstream cells such as horizontal cells (HCs) and amacrine cells (ACs) still aim to create a polychromatic visual perception for survival purposes \citep{osorio2005photoreceptor, dresp2009biological}. The model from \citep{zhang2016retina} proposes two stereotypical pathways for optical signal processing in the retina: a vertical path where signals are collected by photoreceptors, relayed by bipolar cells (BCs), and sent to ganglion cells (GCs), and lateral pathways where local feedbacks transmit information from horizontal cells back to photoreceptors and from amacrine cells to horizontal cells.

Furthermore, cone photoreceptors consist of three types (S-cones, M-cones, and L-cones), each sensitive to different wavelengths, roughly corresponding to the red, green, and blue (RGB) channels of a color image. The study in \citep{zhang2016retina} suggests an overall split-then-combine computational flowchart.

Our model also considers the electrophysiological properties of neurons in the retina. Unlike most neurons in the cerebral cortex, some retinal neurons are so small that local graded potentials can propagate from upstream synapses to downstream somas. For instance, bipolar cells directly excite ganglion cells via local graded potentials from synapses between photoreceptors and bipolar cells. We assume that photoreceptors also have a direct impact on ganglion cells, and we propose a flow of optical signal processing as shown in Fig. 1.

According to \citep{zhang2016retina}, the information processed by HCs can be represented by equation \ref{eq:1}:

\begin{equation} \label{eq:1}
h(x, y)=I_c(x, y) * g(x, y ; \sigma(x, y))
\end{equation}

However, the circuits before BCs also introduce a recursive modulation to channel data \ref{eq:2}, which can cause numerical instability.

\begin{equation} \label{eq:2}
    b(x, y)=\frac{I_c(x, y)}{\alpha+h(x, y)}
\end{equation}

To avoid this issue, we simplify the equation to a more straightforward finite impulse response form as in equation \ref{eq:3}:

\begin{equation} \label{eq:3}
    b(x, y)=\alpha \cdot I_c(x, y)+\beta \cdot I_c(x, y) h(x, y)
\end{equation}

This can be further simplified to a residual form in equation \ref{eq:4}:

\begin{equation} \label{eq:4}
b(x, y)=I_c(x, y)+h(x, y)
\end{equation}

Next, along the optical signal processing neural circuitry, BCs exert a double-opponent effect on the signal modulated by HCs, which can be modeled as a convolution with a difference of Gaussian (DoG) kernel, as shown in equation \ref{eq:5}

\begin{equation} \label{eq:5}
v(x, y)=b(x, y) * f(x, y)
\end{equation}

Although neural networks generally lack the constraint to exert such an effect, we only require the initialization of the corresponding convolutional layer weights to comply with a DoG kernel instance. We choose specific values for this purpose and construct the filter accordingly using equation \ref{eq:6}

\begin{equation} \label{eq:6}
k=G\left(0, \sigma_1\right)-G\left(0, \sigma_2\right)
\end{equation}

As mentioned earlier, bipolar cells relay signals via local potentials rather than action potentials, so we assume that the signals, although attenuated, still potentially exert influence on the ganglion cells. This assumption refines equation (5) to equation \ref{eq:8}:

\begin{equation} \label{eq:8}
v(x, y)=I_c(x, y)+b(x, y) * f(x, y)
\end{equation}

Based on the above computational model, we design the neural network architecture in the next section. This architecture aims to mimic the process of optical signal processing in the retina and provide a more accurate representation of how the retina processes visual information.
The neural network will be designed to handle both the vertical and lateral pathways for optical signal processing, incorporating the various cell types involved in the process. By taking into account the electrophysiological properties of retinal neurons and the specific interactions between photoreceptors, bipolar cells, and ganglion cells, our model offers a more comprehensive and biologically plausible representation of visual processing in the retina.
In conclusion, this computational model strives to accurately represent the complex processing that occurs in the retina, providing valuable insights for the development of improved artificial vision systems and a deeper understanding of the neural mechanisms underlying human vision. By considering the electrophysiological properties of retinal neurons and the interplay between different cell types, this model offers a more complete and biologically plausible representation of the retina's optical signal processing pathways.

\section{Experiment and Results}
\label{sec:experiment}

\subsection{Dataset Utilized}
In our experiment, we employ the open-sourced LOw-Light (LOL) image dataset \citep{wei2018deep}. This dataset comprises 500 pairs of low-light and normal-light images, split into 485 training pairs and 15 testing pairs. The images predominantly feature indoor scenes and have a resolution of 400x600. Given that this image size is comparable to those captured by applications like low-light surveillance, we process the images without down-sampling to represent real-world scenarios.

\subsection{Neural Network Design and Configuration}

The RGB channel values are processed independently. As a result, we utilize depthwise convolutions to maintain this separation \citep{howard2017mobilenets}. The overall network structure aligns with equations \ref{eq:1}, \ref{eq:4}, and \ref{eq:8}. It is worth noting that in equation \ref{eq:1}, the $\sigma$ parameter for convolution $g$ relies on pixel values as the operation moves across the monochromatic input $I_c(x, y)$. To reduce computation overhead, we forego determining $\sigma$ for each location, opting instead to let the network learn the optimal weights during training by specifying the configuration.

\begin{figure}[h]
    \centering
    \includegraphics[width=1.0\textwidth]{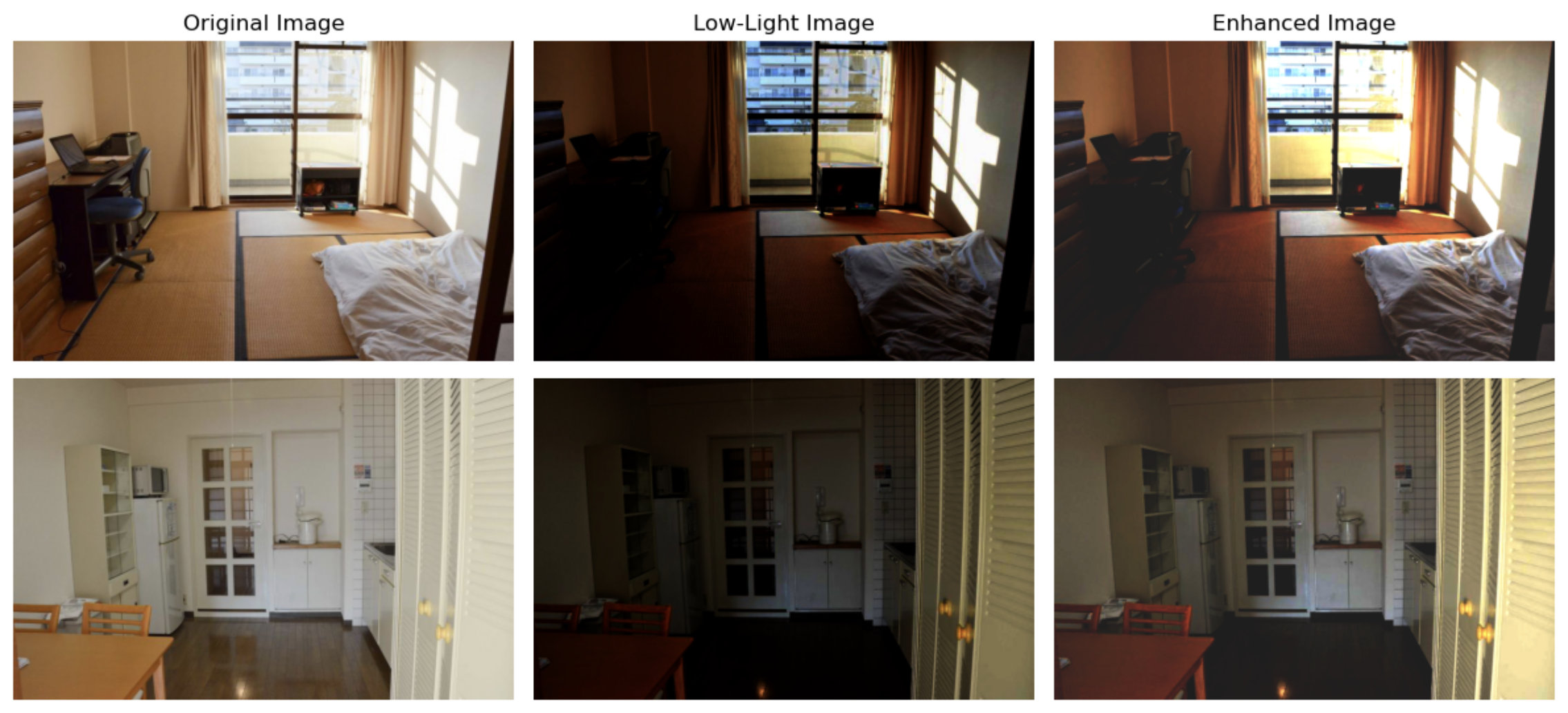}
    \caption{Figure showing (a) Original image, (b) Low-light image, and (c) Enhanced image}
    \label{fig:res}
\end{figure}

Using the above-described configuration, we train the network for 20 epochs with a batch size of 8 and a learning rate of 0.001. The network has only 108 learnable parameters, so the risk of overfitting is minimal, and we do not perform additional validation. Following training, the network is tested directly on the test set. Sample results are shown in Figure \ref{fig:res}, illustrating that our simple network can restore low-light images with a certain level of perceptual quality. Nevertheless the filters' limited size may not effectively extract global illumination information, resulting in darker restored images compared to the ground-truth images.

To evaluate the restored images' quality objectively, we compute the structural similarity index measure (SSIM) for all test samples. The average SSIM is 36.2\%, lower than the 76.3\% to 93.0\% SSIM range reported by other models. Although previous research focused on more theory-oriented heavy models, our straightforward network ensures potential deployment across various systems and scenarios.

\section{Conclusion}
We have developed a simplistic neural network based on the functional principles of retina neurons and applied it to the LIIR problem. By conducting experiments on the benchmark LOL dataset for LIIR tasks, our results indicate a degree of restored image satisfactoriness from a subjective perception standpoint. While the SSIM's objective assessment falls short compared to other methods, the importance of drawing inspiration from biological computation for constructing simple networks is evident, and future work can address the current limitations.

\bibliographystyle{unsrtnat}
\bibliography{paper}

\begin{thebibliography}{29}
\providecommand{\natexlab}[1]{#1}
\providecommand{\url}[1]{\texttt{#1}}
\expandafter\ifx\csname urlstyle\endcsname\relax
  \providecommand{\doi}[1]{doi: #1}\else
  \providecommand{\doi}{doi: \begingroup \urlstyle{rm}\Url}\fi

\bibitem[Gollisch and Meister(2010)]{gollisch2010eye}
Tim Gollisch and Markus Meister.
\newblock Eye smarter than scientists believed: neural computations in circuits
  of the retina.
\newblock \emph{Neuron}, 65\penalty0 (2):\penalty0 150--164, 2010.

\bibitem[Franke et~al.(2017)Franke, Berens, Schubert, Bethge, Euler, and
  Baden]{franke2017inhibition}
Katrin Franke, Philipp Berens, Timm Schubert, Matthias Bethge, Thomas Euler,
  and Tom Baden.
\newblock Inhibition decorrelates visual feature representations in the inner
  retina.
\newblock \emph{Nature}, 542\penalty0 (7642):\penalty0 439--444, 2017.

\bibitem[Liu et~al.(2021)Liu, Hong, Rieke, and Manookin]{liu2021predictive}
Belle Liu, Arthur Hong, Fred Rieke, and Michael~B Manookin.
\newblock Predictive encoding of motion begins in the primate retina.
\newblock \emph{Nature neuroscience}, 24\penalty0 (9):\penalty0 1280--1291,
  2021.

\bibitem[Land and McCann(1971)]{land1971lightness}
Edwin~H Land and John~J McCann.
\newblock Lightness and retinex theory.
\newblock \emph{Josa}, 61\penalty0 (1):\penalty0 1--11, 1971.

\bibitem[Meylan and Susstrunk(2006)]{meylan2006high}
Laurence Meylan and Sabine Susstrunk.
\newblock High dynamic range image rendering with a retinex-based adaptive
  filter.
\newblock \emph{IEEE Transactions on image processing}, 15\penalty0
  (9):\penalty0 2820--2830, 2006.

\bibitem[Wang et~al.(2013)Wang, Zheng, Hu, and Li]{wang2013naturalness}
Shuhang Wang, Jin Zheng, Hai-Miao Hu, and Bo~Li.
\newblock Naturalness preserved enhancement algorithm for non-uniform
  illumination images.
\newblock \emph{IEEE transactions on image processing}, 22\penalty0
  (9):\penalty0 3538--3548, 2013.

\bibitem[Zhang and Li(2016)]{zhang2016retina}
Xian-Shi Zhang and Yong-Jie Li.
\newblock A retina inspired model for high dynamic range image rendering.
\newblock In \emph{Advances in Brain Inspired Cognitive Systems: 8th
  International Conference, BICS 2016, Beijing, China, November 28-30, 2016,
  Proceedings 8}, pages 68--79. Springer, 2016.

\bibitem[Park et~al.(2017)Park, Yu, Moon, Ko, and Paik]{park2017low}
Seonhee Park, Soohwan Yu, Byeongho Moon, Seungyong Ko, and Joonki Paik.
\newblock Low-light image enhancement using variational optimization-based
  retinex model.
\newblock \emph{IEEE Transactions on Consumer Electronics}, 63\penalty0
  (2):\penalty0 178--184, 2017.

\bibitem[Chen et~al.(2018)Chen, Chen, Xu, and Koltun]{chen2018learning}
Chen Chen, Qifeng Chen, Jia Xu, and Vladlen Koltun.
\newblock Learning to see in the dark.
\newblock In \emph{Proceedings of the IEEE conference on computer vision and
  pattern recognition}, pages 3291--3300, 2018.

\bibitem[Wang et~al.(2020{\natexlab{a}})Wang, Chen, and Hoi]{wang2020deep}
Zhihao Wang, Jian Chen, and Steven~CH Hoi.
\newblock Deep learning for image super-resolution: A survey.
\newblock \emph{IEEE transactions on pattern analysis and machine
  intelligence}, 43\penalty0 (10):\penalty0 3365--3387, 2020{\natexlab{a}}.

\bibitem[Ma et~al.(2021)Ma, Mao, Zheng, Gao, Wang, Han, Ding, Zhang, and
  Doermann]{ma2021oriented}
Teli Ma, Mingyuan Mao, Honghui Zheng, Peng Gao, Xiaodi Wang, Shumin Han, Errui
  Ding, Baochang Zhang, and David Doermann.
\newblock Oriented object detection with transformer.
\newblock \emph{arXiv preprint arXiv:2106.03146}, 2021.

\bibitem[Wang et~al.(2021)Wang, Zhao, Dou, Yu, Liu, and Wu]{wang2021compact}
Shuyu Wang, Mingxin Zhao, Runjiang Dou, Shuangming Yu, Liyuan Liu, and Nanjian
  Wu.
\newblock A compact high-quality image demosaicking neural network for
  edge-computing devices.
\newblock \emph{Sensors}, 21\penalty0 (9):\penalty0 3265, 2021.

\bibitem[Li et~al.(2019)Li, Miao, Zhang, and Wang]{li2019denoisingnet}
Yang Li, Zhuang Miao, Rui Zhang, and Jiabao Wang.
\newblock Denoisingnet: An efficient convolutional neural network for image
  denoising.
\newblock In \emph{2019 2nd International Conference on Artificial Intelligence
  and Big Data (ICAIBD)}, pages 409--413. IEEE, 2019.

\bibitem[Wu et~al.(2023)Wu, Shi, Shen, Tan, and Wang]{wu2023light}
Yun Wu, Yucheng Shi, Huaiyan Shen, Yaya Tan, and Yu~Wang.
\newblock Light-tbfnet: Rgb-d salient detection based on a lightweight
  two-branch fusion strategy.
\newblock \emph{Multimedia Tools and Applications}, pages 1--31, 2023.

\bibitem[Wang et~al.(2020{\natexlab{b}})Wang, Hong, Wang, and Yu]{wang2020gan}
Wanwei Wang, Wei Hong, Feng Wang, and Jinke Yu.
\newblock Gan-knowledge distillation for one-stage object detection.
\newblock \emph{IEEE Access}, 8:\penalty0 60719--60727, 2020{\natexlab{b}}.

\bibitem[Wei et~al.(2022)Wei, Tian, Wei, Zhong, Chen, Pu, and
  Lu]{wei2022efficient}
Dafeng Wei, Ye~Tian, Liqing Wei, Hong Zhong, Siqian Chen, Shiliang Pu, and
  Hongtao Lu.
\newblock Efficient dual attention slowfast networks for video action
  recognition.
\newblock \emph{Computer Vision and Image Understanding}, 222:\penalty0 103484,
  2022.

\bibitem[Hira et~al.(2021)Hira, Das, Modi, and Pakhomov]{hira2021delta}
Sanchit Hira, Ritwik Das, Abhinav Modi, and Daniil Pakhomov.
\newblock Delta sampling r-bert for limited data and low-light action
  recognition.
\newblock In \emph{Proceedings of the IEEE/CVF Conference on Computer Vision
  and Pattern Recognition}, pages 853--862, 2021.

\bibitem[Ren et~al.(2020)Ren, Yang, Cheng, and Liu]{ren2020lr3m}
Xutong Ren, Wenhan Yang, Wen-Huang Cheng, and Jiaying Liu.
\newblock Lr3m: Robust low-light enhancement via low-rank regularized retinex
  model.
\newblock \emph{IEEE Transactions on Image Processing}, 29:\penalty0
  5862--5876, 2020.

\bibitem[Zhang and Wu(2016)]{zhang2016fast}
Xin Zhang and Ruiyuan Wu.
\newblock Fast depth image denoising and enhancement using a deep convolutional
  network.
\newblock In \emph{2016 IEEE International Conference on Acoustics, Speech and
  Signal Processing (ICASSP)}, pages 2499--2503. IEEE, 2016.

\bibitem[Tian et~al.(2020)Tian, Xu, Li, Zuo, Fei, and Liu]{tian2020attention}
Chunwei Tian, Yong Xu, Zuoyong Li, Wangmeng Zuo, Lunke Fei, and Hong Liu.
\newblock Attention-guided cnn for image denoising.
\newblock \emph{Neural Networks}, 124:\penalty0 117--129, 2020.

\bibitem[Hassan et~al.(2022)Hassan, Mishra, Ahmad, Bashir, Huang, and
  Luo]{hassan2022effects}
Haseeb Hassan, Pranshu Mishra, Muhammad Ahmad, Ali~Kashif Bashir, Bingding
  Huang, and Bin Luo.
\newblock Effects of haze and dehazing on deep learning-based vision models.
\newblock \emph{Applied Intelligence}, pages 1--19, 2022.

\bibitem[Zhang et~al.(2021)Zhang, Guo, Huang, and Wen]{zhang2021rellie}
Rongkai Zhang, Lanqing Guo, Siyu Huang, and Bihan Wen.
\newblock Rellie: Deep reinforcement learning for customized low-light image
  enhancement.
\newblock In \emph{Proceedings of the 29th ACM international conference on
  multimedia}, pages 2429--2437, 2021.

\bibitem[Jing et~al.(2020)Jing, Liu, Ding, Wang, Ding, Song, and
  Wen]{jing2020dynamic}
Yongcheng Jing, Xiao Liu, Yukang Ding, Xinchao Wang, Errui Ding, Mingli Song,
  and Shilei Wen.
\newblock Dynamic instance normalization for arbitrary style transfer.
\newblock In \emph{Proceedings of the AAAI Conference on Artificial
  Intelligence}, volume~34, pages 4369--4376, 2020.

\bibitem[Jiang et~al.(2021)Jiang, Gong, Liu, Cheng, Fang, Shen, Yang, Zhou, and
  Wang]{jiang2021enlightengan}
Yifan Jiang, Xinyu Gong, Ding Liu, Yu~Cheng, Chen Fang, Xiaohui Shen, Jianchao
  Yang, Pan Zhou, and Zhangyang Wang.
\newblock Enlightengan: Deep light enhancement without paired supervision.
\newblock \emph{IEEE transactions on image processing}, 30:\penalty0
  2340--2349, 2021.

\bibitem[Zheng and Gupta(2022)]{zheng2022semantic}
Shen Zheng and Gaurav Gupta.
\newblock Semantic-guided zero-shot learning for low-light image/video
  enhancement.
\newblock In \emph{Proceedings of the IEEE/CVF Winter Conference on
  Applications of Computer Vision}, pages 581--590, 2022.

\bibitem[Osorio and Vorobyev(2005)]{osorio2005photoreceptor}
Daniel Osorio and Misha Vorobyev.
\newblock Photoreceptor sectral sensitivities in terrestrial animals:
  adaptations for luminance and colour vision.
\newblock \emph{Proceedings of the Royal Society B: Biological Sciences},
  272\penalty0 (1574):\penalty0 1745--1752, 2005.

\bibitem[Dresp-Langley and Langley(2009)]{dresp2009biological}
Birgitta Dresp-Langley and Keith Langley.
\newblock The biological significance of colour perception.
\newblock \emph{Color perception: Physiology, processes and analysis}, page~89,
  2009.

\bibitem[Wei et~al.(2018)Wei, Wang, Yang, and Liu]{wei2018deep}
Chen Wei, Wenjing Wang, Wenhan Yang, and Jiaying Liu.
\newblock Deep retinex decomposition for low-light enhancement.
\newblock \emph{arXiv preprint arXiv:1808.04560}, 2018.

\bibitem[Howard et~al.(2017)Howard, Zhu, Chen, Kalenichenko, Wang, Weyand,
  Andreetto, and Adam]{howard2017mobilenets}
Andrew~G Howard, Menglong Zhu, Bo~Chen, Dmitry Kalenichenko, Weijun Wang,
  Tobias Weyand, Marco Andreetto, and Hartwig Adam.
\newblock Mobilenets: Efficient convolutional neural networks for mobile vision
  applications.
\newblock \emph{arXiv preprint arXiv:1704.04861}, 2017.

\end{thebibliography}
\end{document}